\documentclass[conference]{IEEEtran}
\makeatletter
\let\NAT@parse\undefined
\makeatother
\usepackage{times}
\usepackage{siunitx} 
\usepackage{multicol} 
\usepackage[bookmarks=true]{hyperref} 
\usepackage{cuted} 
\usepackage{caption} 
\usepackage{kotex} 
\usepackage{booktabs} 
\usepackage{graphicx} 
\usepackage{float} 
\usepackage[table]{xcolor} 
\usepackage{multirow} 
\usepackage{amsmath,amssymb}

\usepackage{tabularx} 
\usepackage{makecell} 
\usepackage{placeins} 
\usepackage{times} 
\usepackage[most]{tcolorbox}
 
\usepackage{color} 
\usepackage{soul} 
\begin{document}

\title{Selective Perception for Robot:\\
Task-Aware Attention in Multimodal VLA}

\author{
\IEEEauthorblockN{Young-Chae Son$^{\ddagger}$, Jung-Woo Lee$^{\ddagger}$,
Yoon-Ji Choi, Dae-Kwan Ko, Soo-Chul Lim*}

\IEEEauthorblockA{
Dongguk University
}
}

\maketitle
\begin{strip}
    \centering
    \vspace*{-1.3cm}
    \includegraphics[width=\textwidth]{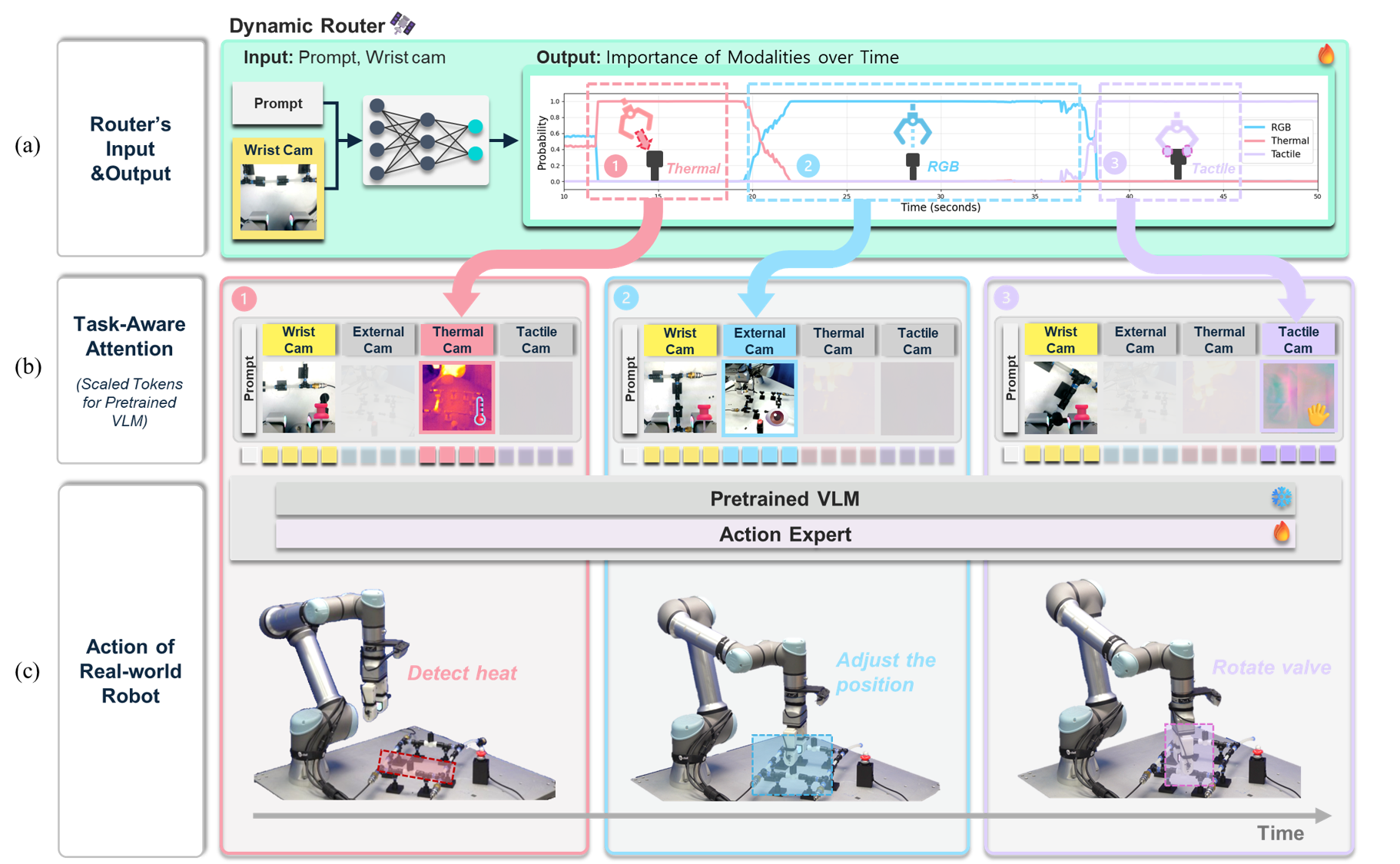}
\captionof{figure}{\textbf{Operational process of the proposed model.} 
(a) \textbf{Camera Router Module:} The router utilizes the wrist camera and text prompt to analyze the current context and predict the real-time importance of each modality. 
(b) \textbf{Task-Aware Attention:} Based on the router's prediction, the model selectively attends to the most informative visual streams.
(c) \textbf{Real-world Execution:} Robot action sequence for Task 3 (\textit{Rotate the valve on the hot side and hold down the front button}).}
\end{strip}

\begin{abstract}
In robotics, Vision-Language-Action (VLA) models that integrate diverse multimodal signals from multi-view inputs have emerged as an effective approach. However, most prior work adopts static fusion that processes all visual inputs uniformly, which incurs unnecessary computational overhead and allows task-irrelevant background information to act as noise. Inspired by the principles of human active perception, we propose a dynamic information fusion framework designed to maximize the efficiency and robustness of VLA models. Our approach introduces a lightweight adaptive routing architecture that analyzes the current text prompt and observations from a wrist-mounted camera in real-time to predict the task-relevance of multiple camera views. By conditionally attenuating computations for views with low informational utility and selectively providing only essential visual features to the policy network, Our framework achieves computation efficiency proportional to task relevance. Furthermore, to efficiently secure large-scale annotation data for router training, we established an automated labeling pipeline utilizing Vision-Language Models (VLMs) to minimize data collection and annotation costs. Experimental results in real-world robotic manipulation scenarios demonstrate that the proposed approach achieves significant improvements in both inference efficiency and control performance compared to existing VLA models, validating the effectiveness and practicality of dynamic information fusion in resource-constrained, real-time robot control environments.
\end{abstract}

\IEEEpeerreviewmaketitle

\section{Introduction}
To realize general-purpose robotic intelligence that can flexibly adapt to diverse task environments, research on Vision–Language–Action (VLA) models that integrate visual, linguistic, and action-related information has been accelerating~\cite{kim2025openvla,octo_2023,black2024pi_0,qu2025spatialvla,sapkota2025visionlanguageactionmodelsconceptsprogress}. While early VLA models predominantly relied on RGB images as their primary visual input~\cite{brohan2022rt,zitkovich2023rt,kim2025openvla},this approach often exhibits limitations in real-world environments. Consequently, recent research has expanded to incorporate diverse sensory modalities, such as depth information~\cite{li2025pointvla,zhen20243d,zawalski2025robotic}, tactile signals~\cite{huang2025tactile,yu2025forcevla,hao2025tla,zhang2025vtla}, and audio cues~\cite{zhao2025vlas,liu2025maniwav}, into the model. Such research on multimodal integration~\cite{jones24fuse} demonstrates that specialized tasks, previously unachievable by existing VLAs, can be performed while preserving large-scale prior knowledge. 

However, these approaches remain limited to static fusion, resulting in only an expansion of input modalities. Such static fusion suffers from two fundamental limitations. First, it gives rise to issues of information redundancy and interference. When irrelevant data are indiscriminately projected into high-dimensional feature spaces, it hinders effective representation learning and degrades both training efficiency and convergence speed. Second, it can result in a degradation of robustness. At certain stages, irrelevant modalities can act as noise that dilutes meaningful information, negatively affecting the model's reliability and generalizability.

The importance of sensory information varies according to the type of tasks in real-world environments. To address these challenges, the human cognitive system employs active perception which actively modulates the weights of sensory information based on task contexts and goals~\cite{birman2019flexible}. This demonstrates that multimodal sensory processing involves not only integration but also inhibitory interactions across modalities~\cite{ide2013tactile,ide2016neural,hidaka2015sound}. This suggests that human perception does not indiscriminately process all sensory inputs; rather, it leverages attention to suppress irrelevant signals and selectively allocate resources to information most relevant to the current task. For example, in \textit{Turn off an overheated electronic device} task, humans rely on visual localization cues during the initial exploration phase, shift their attention to thermal sensing when assessing potential risk, and prioritize tactile feedback during object grasping. During this process, other sensory inputs are temporarily suppressed or attenuated. This implies that humans dynamically select which sensory modality to trust based on the task phase and objective, thereby minimizing interference from irrelevant information.

Inspired by the selective inhibition mechanisms observed in human perception, we propose a Dynamic Routing-based adaptive multimodal fusion approach that dynamically suppresses task-irrelevant sensory data rather than processing all inputs uniformly. In contrast to prior approaches that rely on expert-defined rules or manually segmented phase labels, we introduce a Vision–Language Model (VLM)–based auto-labeling pipeline that enables the generation of supervisory signals for large-scale datasets without human intervention. Furthermore, the proposed router is integrated with a Flow Matching–based VLA policy network, enabling it to be trained during robot fine-tuning without the need for additional training data or labeling. Consequently, the proposed framework avoids indiscriminate information fusion by prioritizing the integration of the most relevant modalities at each time step, thereby minimizing information interference in high-dimensional feature spaces and improving both task accuracy and efficiency.

\noindent \textbf{The key contributions of this work are summarized as follows:}
\begin{itemize}
  \item \textbf{Context-Aware Router \& Gating:} We propose a Dynamic Router that uses the wrist-mounted camera as an anchor to predict modality importance in real-time.
  
  \item \textbf{Robustness in Long-Horizon Tasks:} By leveraging a router-based feature fusion structure to integrate key modalities, we mitigate inter-modality interference and achieve higher success rates than existing methods on long-horizon manipulation tasks.
  
 \item \textbf{Scalable VLM Supervision:} We construct a VLM-based auto-labeling pipeline to obtain labeled data for router training, generating camera labels without additional data collection costs.

\end{itemize}

\section{Related Works}
\begin{figure*}[t]
    \centering
    \includegraphics[scale=0.5]{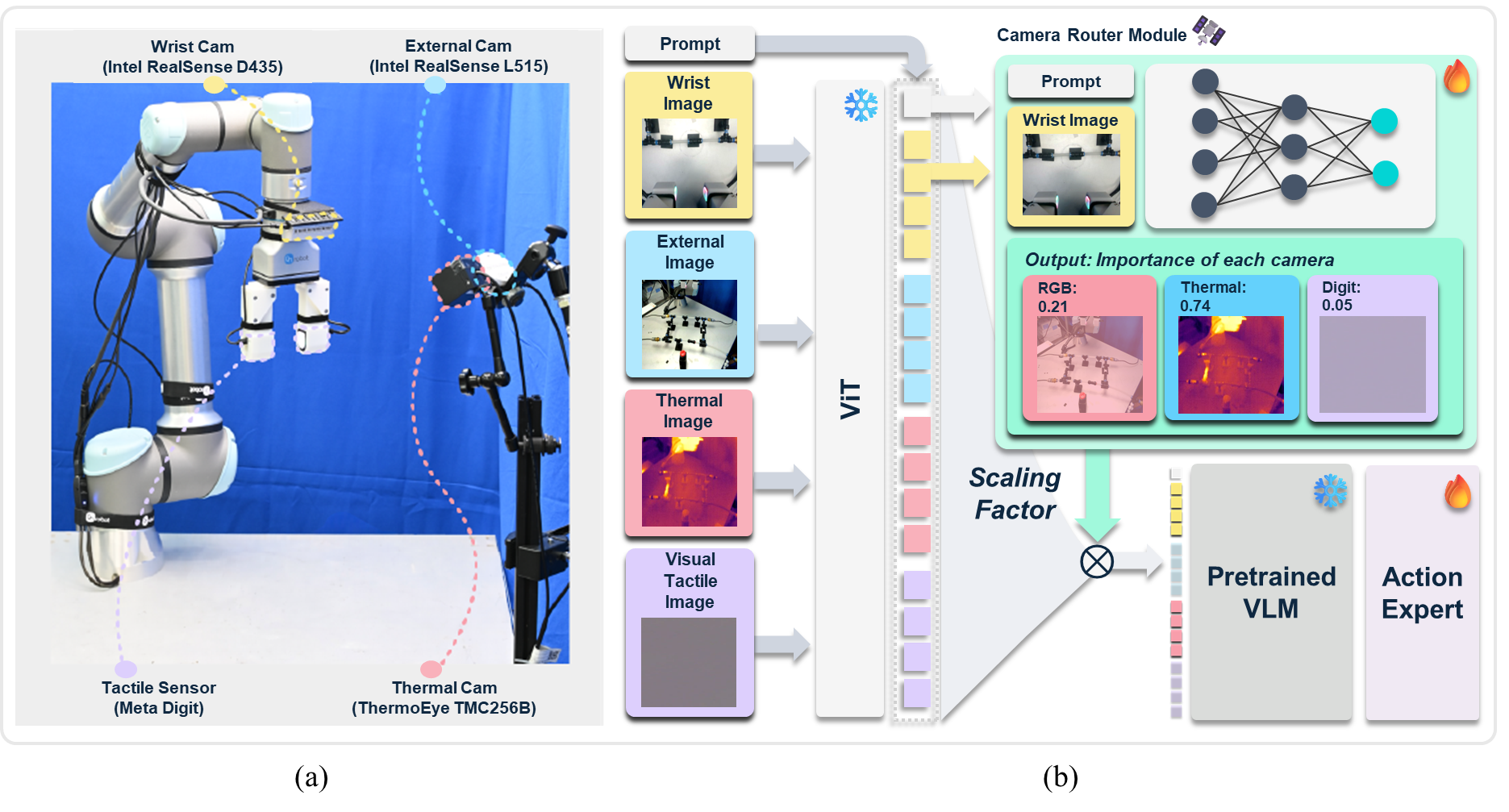} 
    \caption{\textbf{Overview of the proposed framework.} (a) Experiment Setup (b) Architecture of the proposed framework and the dynamic router. The router takes the wrist camera and langage prompt feed as inputs to predict importance weights for multiple visual sources, including external, thermal, and tactile cameras, thereby prioritizing task-relevant sensor information.}
    \label{fig:proposed_framework} 
\end{figure*}

\subsection{Vision-Language-Action (VLA) Models for Robotics}
Vision-Language-Action (VLA) models, which integrally understand vision, language, and action information, enable robots to perform general-purpose tasks by simultaneously comprehending visual context and language instructions~\cite{kim2025openvla,octo_2023,black2024pi_0}. However, complex interactions in real-world environments cannot be fully captured using only 2D image information. Previous studies have improved success rates on specific tasks by leveraging additional sensor modalities~\cite{ko2023vision,lee2024dextouch,10342363,sun2024interactive,10801833,liang2024survey}. For instance, the integration of tactile information is essential for tasks requiring precise contact, such as peg-in-hole tasks. Although TLA~\cite{hao2025tla} demonstrated an 85\% success rate using tactile information alone, the absence of visual feedback limited its ability to capture the global context of the task. Addressing this limitation, VTLA~\cite{zhang2025vtla} introduced the Vision-Tactile-Action-Prompt dataset fusing visual and tactile inputs, leading to significantly more refined control performance.
Furthermore, research has extended to leverage audio information. ManiWAV~\cite{liu2025maniwav} demonstrated that auditory data, such as contact sounds or the sound of liquid flow, can serve as key cues to compensate for the uncertainty of visual information. Multimodal approaches have shown promising results, but most existing models still rely on static fusion strategies that process all incoming sensory data with fixed weights. This approach fails to account for the relative importance of information based on the task context and can lead to information interference due to the inclusion of unnecessary modalities.
 
\subsection{Adaptive Sensory Fusion and Modality Weighting}
For complex tasks in real-world environments, the relative importance of each sensory modality varies dynamically with the task context~\cite{ICLR2025_e1126028,feng2025play,he2025foar}. As a result, research has investigated adaptively controlling modality-specific weights according to task stages.
With respect to visual information, prior work has explored predicting feature importance for each camera view in multi-view settings to reweight and fuse features~\cite{11080054}. In addition, Long-VLA~\cite{fan2025longvlaunleashinglonghorizoncapability} demonstrated that masking unnecessary visual information based on the task stage can enhance success rates in long-horizon tasks.

Recent studies have also explored approaches that leverage physical properties. Play to the Score~\cite{feng2025play} proposed a method that dynamically fuses diverse sensory information by adjusting modality weights in real-time, in a manner analogous to human perception. This approach demonstrated the capability to autonomously learn and determine when and which information to prioritize, rather than being limited to simple multimodal integration. FOAR~\cite{he2025foar} introduced a structural gating approach that modulates the weight of force/torque data based on contact status, thereby effectively mitigating noise interference in non-contact scenarios. However, this approach relied on manual annotation by humans. To address this limitation, subsequent studies proposed dynamically fusing visual and tactile modalities through attention-based architectures designed to predict future contact forces~\cite{li2025adaptive}. Despite these efforts, research on dynamically adjusting weights for each sensory modality in VLA frameworks remains limited.
To address this, we propose leveraging the context-aware capabilities of VLMs to explicitly define modalities optimized for the task context and train the model to learn task-relevant sensory information.

\subsection{Scalable Supervision and Automated Labeling}
Traditional data curation for robotic learning relies on manual human annotation. However, this approach is costly and prone to inconsistency due to annotator fatigue or varying levels of expertise~\cite{li2025adaptive,yu2025rlaif}. Recent studies have demonstrated that Large Language Models (LLMs) and Vision Language Models (VLMs) can perform diverse tasks by leveraging their extensive prior knowledge and high-level reasoning capabilities~\cite{kwon2024language,singh2024malmm}. In particular, the visual reasoning capabilities of VLMs have been shown to effectively support robotic task execution~\cite{mei_replanvlm_2024}. Building upon these findings, recent studies have shown that VLM-based automated labeling can serve as a viable alternative for overcoming these limitations~\cite{chen2024sharegpt4v,wang2024robogen}. For example, AutoRT~\cite{ahn2024autort} demonstrated that leveraging VLMs enables scene understanding and reasoning, thereby effectively scaling up the size of training dataset.
Adopting this approach, we utilize VLMs to alleviate the burden of manual annotation in constructing large-scale datasets. Specifically, our approach designs the VLM to automatically assign relative importance weights to key multimodal features. This enables the Router, trained alongside a VLA model, to make real-time decisions on sensory modality prioritization while performing tasks.

\section{Method}
In this section, we propose a VLA model architecture that operates effectively in multi-view environments. To achieve this, we introduce a Camera Router between the ViT and the pretrained VLM. This module computes the importance of the input information from each viewpoint. This importance ensures that visual information provided to the policy network is proportional to its relevance for task execution. This section is organized as follows: Section A presents the integrated training process of the Router and the VLA model, Section B details the architecture of the Camera Router, and Section C outlines the methodology for constructing a dataset for Router training using VLMs.

\begin{figure*}[t]
    \includegraphics[width=\textwidth]{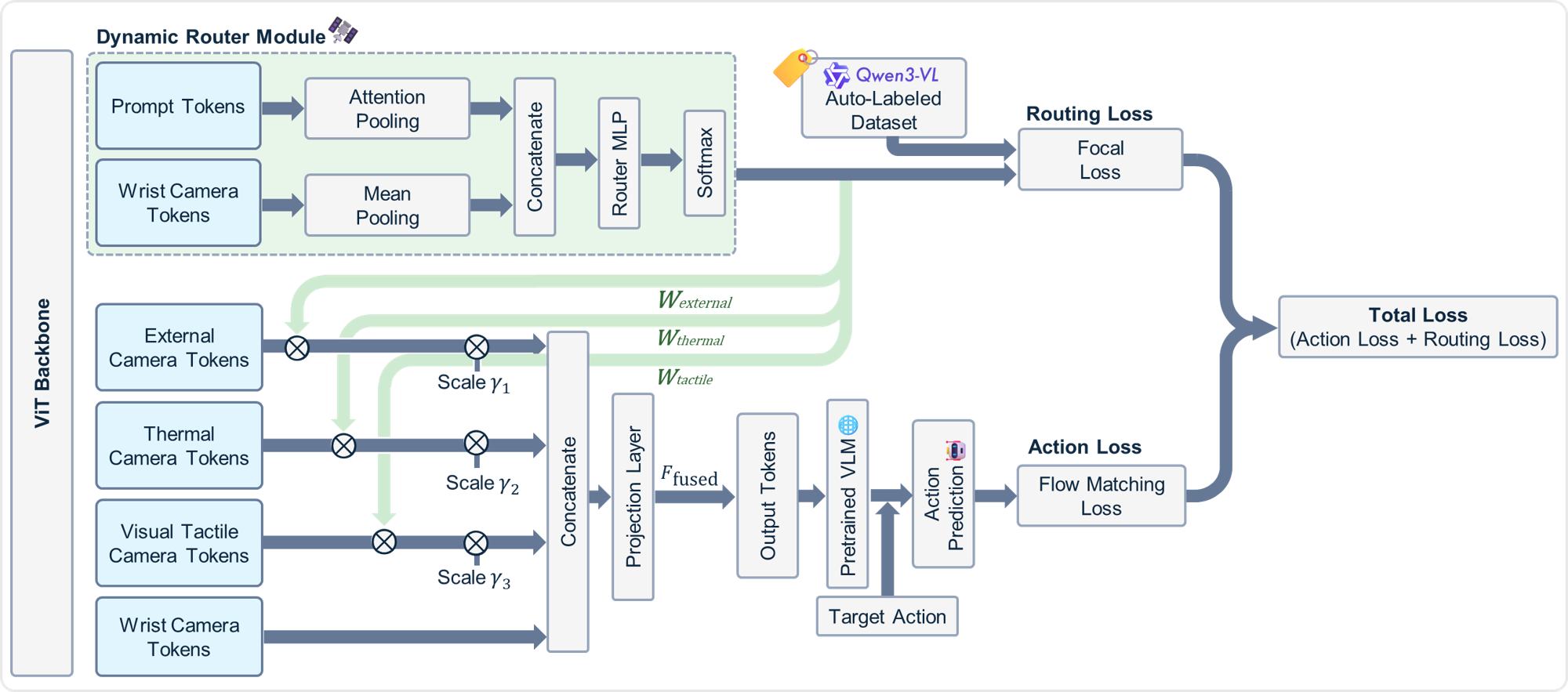} 
    \caption{\textbf{Overview of VLA Training with Dynamic Router.} The Camera Router predicts importance weights ($W$) for multiple camera views, acting as a gating mechanism. This allows the VLA framework to selectively prioritize task-relevant visual information during the action training process.}
    \label{fig:router_architecture} 
\end{figure*}

\subsection{Training with Vision Language Action Model}
Fig.~\ref{fig:proposed_framework} illustrates the training process for both the VLA and the router. Full fine-tuning of large-scale VLA models for specific tasks requires substantial computational resources and large-scale training data. To address these computational and data efficiency challenges, we adopt a fine-tuning strategy that applies LoRA to the pretrained base policy~\cite{black2024pi_0}. This approach keeps the billions of parameters in the pretrained base policy frozen. Instead, we insert lightweight, trainable LoRA modules into all weight groups of both the VLM and the Action expert, and train only these specific modules using our collected dataset. This approach enables us to perform the desired task actions using only 50 episodes and minimal computational resources, while preserving the extensive prior knowledge of base policy.
The proposed framework is trained in an end-to-end manner to simultaneously achieve optimal robotic action generation and efficient camera routing. The overall loss function $L_{\mathrm{total}}$ is defined as a weighted sum of the action loss for action prediction and the routing loss for router optimization:

\begin{equation}
L_{\mathrm{total}} = L_{\mathrm{action}} + \lambda L_{\mathrm{routing}}
\end{equation}

where $\lambda$ represents the weight for routing loss function, which is set to 0.1 to ensure effective routing performance without degrading the primary action prediction objective.

We use Flow Matching Loss for $L_{\mathrm{action}}$, in accordance with the training strategy of the base policy. During training, the Camera Router is supervised using $L_{\mathrm{routing}}$, while simultaneously being optimized to improve the VLA's action prediction performance through gradient backpropagation from $L_{\mathrm{action}}$.
Finally, the router weight and the projection layer undergo full parameter updates, whereas only the LoRA parameters are updated for the VLA backbone, thereby maximizing training efficiency.

\subsection{Camera Router Architecture}
Fig.~\ref{fig:router_architecture} illustrates the overall architecture of the proposed router. To perceive the current context, the router takes the wrist camera image and the task prompt as inputs. The detailed feature extraction process proceeds as follows. First, We apply Attention Pooling to the task prompt to obtain the text feature $h_{\mathrm{prompt}}$ that captures core keywords. In parallel, the wrist camera tokens are transformed into the visual feature $h_{\mathrm{wrist}}$ through Mean Pooling. In this paper, the wrist camera is designated as the base anchor view and designed to remain always active. This design ensures the stable availability of proximal visual information, which is considered essential for robotic manipulation.

\begin{equation}
W = \mathrm{Softmax}(\mathrm{MLP}(x)) \in \mathbb{R}^{K-1}
\end{equation}

Finally, the router input $x$ is defined as the concatenation of the prompt and wrist camera features, which is then passed through an MLP and a Softmax layer to predict the importance weights $W$ for the remaining $K$ $-$ $1$ views.

\begin{figure*}[t]
    \centering
    \includegraphics[width=\textwidth]{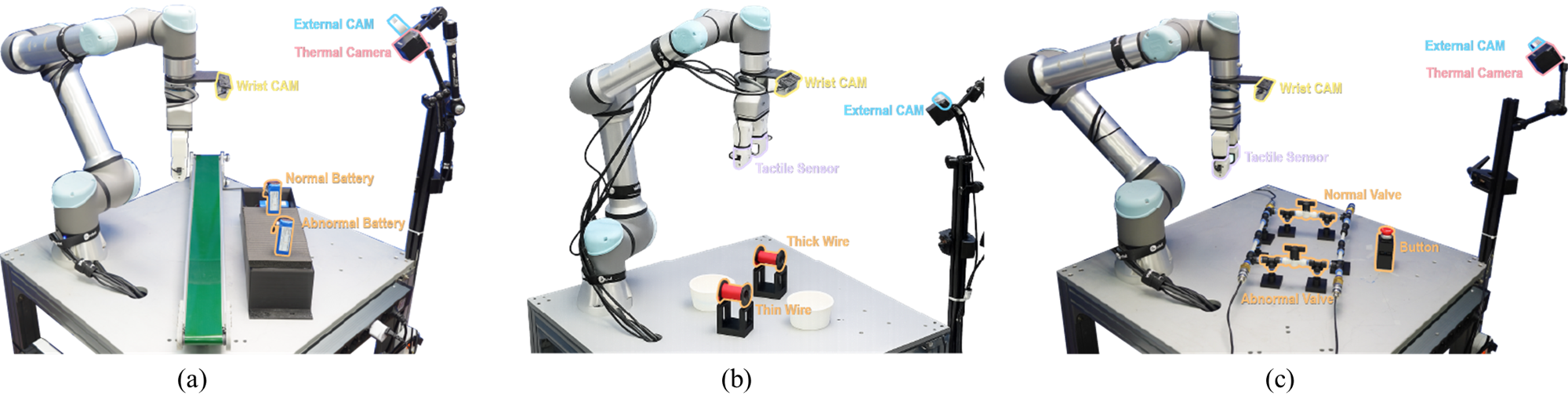} 
\caption{\textbf{Experimental Setup for Each Task.} (a) Battery Sorting, (b) Wire Selection, and (c) Valve Operation. While all tasks share a dual RGB camera system for visual perception, each setup incorporates task-specific modalities—such as thermal cameras or tactile sensors—to provide complementary sensory information.}
    \label{fig:experiment_setup} 
\end{figure*}

\begin{equation}
W = [W_{\text{external}}, W_{\text{thermal}}, W_{\text{tactile}}]
\end{equation}

\subsubsection{Dynamic Gating}

The Dynamic Router predicts the weights $w$, which determine the contribution 
of each modality to the ongoing task and are applied in conjunction with a 
learnable parameter $\gamma \in \mathbb{R}^{D}$. 
\begin{equation}
F_{\mathrm{gated}} = \gamma \odot (w \odot F_{\mathrm{original}})
\end{equation}

Here, $\gamma$ is a scaling parameter for mitigating abrupt changes during the gating process and enhancing training stability.
Through this process, features from  camera views with lower importance are naturally attenuated, while task-relevant visual information is relatively emphasized. Subsequently, the visual features are concatenated along the channel dimension and fused into a unified visual representation. Specifically, the wrist camera feature $F_{\mathrm{wrist}}$ is concatenated with the gated external camera and multimodal features $F_{\mathrm{external}}'$, $F_{\mathrm{thermal}}'$, and $F_{\mathrm{tactile}}'$. The concatenated features are then projected back to the original embedding dimension using a linear projection. Finally, the fused feature $F_{\mathrm{fused}}$ is used as the input for the pretrained VLM, providing the model with visual information that is selectively emphasized based on the task context.

\subsubsection{Imbalanced Routing Learning}

In robotic manipulation tasks, certain sensory modalities (e.g. tactile, thermal) are not required throughout the entire task, but are sparsely needed only at specific moments involving contact or decision-making. Due to this data imbalance, the use of a standard cross-entropy loss may lead to biased model training. To address the imbalance of sparsely occurring ground-truth labels, we train the router using Focal Loss.

\begin{equation}
    L_{\mathrm{routing}} = -\alpha_{\text{focal}} (1 - p_t)^{\beta} \log(p_t)
\end{equation}

Focal Loss reduces the influence of easy samples, thereby reweighting the learning focus toward difficult and sparsely occurring instances. This approach allows for effective training by preventing the router from missing sparse, high-value moments. We set $\alpha_{\text{focal}}=0.25$ and $\beta=2.0$.

\subsection{Auto-Labeling with VLM Labels}
Task decomposition has been explored in robotics, as such structured approaches~\cite{triantafyllidis2023hybrid,dalal2025local} are known to enhance task performance. However, manually annotating task stages or required sensory data for every frame in large-scale datasets is prohibitively time-consuming and expensive, posing a bottleneck to scalability. To address this challenge, we established an automated pipeline utilizing Qwen3-VL~\cite{Qwen3-VL} to recognize task phases within videos and automatically generate the corresponding camera importance labels. Determining the robot's state from a single frame alone is inherently ambiguous. For example, a static grasping state can be difficult to interpret as either the onset of Transporting or Placing. To overcome this limitation, we employ a History-Aware Prompting strategy that incorporates past action history in addition to the current visual information. The collected input videos are segmented into chunks of $T=30$ frames. To capture the temporal context within each chunk, we uniformly sample $K=10$ images, which are fed into the VLM. To facilitate context understanding, the prompt is constructed using the following key elements:

\begin{enumerate}
  \item \emph{Role:} The assigned persona defining the agent's behavioral context.
  \item \emph{Task Description:} The overall objective of the task.
  \item \emph{Workflow Definition:} The sequence of actions during task execution.
  \item \emph{Gripper Action:} The scalar value representing the gripper's opening width.
  \item \emph{Action History:} A summary of the previous three chunks.
\end{enumerate}

Through this process, the VLM jointly considers the visual inputs and contextual information to determine the activation status of each camera according to predefined rules: 
\begin{equation}
    D_{\mathrm{vlm}} = [c_{\mathrm{ext}}, c_{\mathrm{thm}}, c_{\mathrm{tac}}]^\top \in \{0, 1\}^3
\end{equation}

The generated $D_{\mathrm{vlm}}$ provides supervision for the Router through the Focal Loss described in Section B, enabling it to learn critical segments.

\section{Experiment}

 \begin{table*}[t]
    \centering
    \caption{\textbf{Quantitative comparison of subtask success rates.} Bold values indicate the best performance in each row.}
    \label{tab:success_rate_symmetric_clean}
    \setlength{\aboverulesep}{0.5pt}
    \setlength{\belowrulesep}{0.5pt}
    \renewcommand{\arraystretch}{1.5} 
    
    \begin{tabular*}{\textwidth}{@{\extracolsep{\fill}}lcclccc}

        \cmidrule(lr){1-3} \cmidrule(lr){4-7} 
        % 1. 최상단 그룹 헤더
         & \multicolumn{2}{c}{\textbf{w/o Router}} & & \multicolumn{3}{c}{\textbf{w/ Router}} \\
        \cmidrule(lr){2-3} \cmidrule(lr){5-7} 
        
        % 2. 서브 헤더
        \textbf{Policy Input} & \textbf{w/o Multimodal} & \textbf{w/ Multimodal} & \textbf{Router Input} & \textbf{Wrist Cam + State} & \textbf{Ext. Cam + Prompt} & \textbf{Wrist Cam + Prompt} \\
        \cmidrule(lr){1-3} \cmidrule(lr){4-7} 
        
        \textbf{Task 1} & 
        \begin{tabular}[c]{@{}c@{}} 0.00\% \\[-3pt] (0/10) \end{tabular} & 
        \begin{tabular}[c]{@{}c@{}} 30.00\% \\[-3pt] (3/10) \end{tabular} & 
        \textbf{Task 1} & 
        \begin{tabular}[c]{@{}c@{}} 20.00\% \\[-3pt] (2/10) \end{tabular} & 
        \begin{tabular}[c]{@{}c@{}} \textbf{90.00\%} \\[-3pt] \textbf{(9/10)} \end{tabular} & 
        \begin{tabular}[c]{@{}c@{}} \textbf{90.00\%} \\[-3pt] \textbf{(9/10)} \end{tabular} \\
        
        \textbf{Task 2} & 
        \begin{tabular}[c]{@{}c@{}} 0.00\% \\[-3pt] (0/10) \end{tabular} & 
        \begin{tabular}[c]{@{}c@{}} 0.00\% \\[-3pt] (0/10) \end{tabular} & 
        \textbf{Task 2} & 
        \begin{tabular}[c]{@{}c@{}} 60.00\% \\[-3pt] (6/10) \end{tabular} & 
        \begin{tabular}[c]{@{}c@{}} 70.00\% \\[-3pt] (7/10) \end{tabular} & 
        \begin{tabular}[c]{@{}c@{}} \textbf{90.00\%} \\[-3pt] \textbf{(9/10)} \end{tabular} \\
        
        \textbf{Task 3} & 
        \begin{tabular}[c]{@{}c@{}} 0.00\% \\[-3pt] (0/10) \end{tabular} & 
        \begin{tabular}[c]{@{}c@{}} 0.00\% \\[-3pt] (0/10) \end{tabular} & 
        \textbf{Task 3} & 
        \begin{tabular}[c]{@{}c@{}} 10.00\% \\[-3pt] (1/10) \end{tabular} & 
        \begin{tabular}[c]{@{}c@{}} 60.00\% \\[-3pt] (6/10) \end{tabular} & 
        \begin{tabular}[c]{@{}c@{}} \textbf{70.00\%} \\[-3pt] \textbf{(7/10)} \end{tabular} \\
        
        \cmidrule(lr){1-3} \cmidrule(lr){4-7} 
        \textbf{Average SR} & \boldmath{$0.00 \pm 0.00\%$} & \boldmath{$10.00 \pm 17.32\%$} & 
        \textbf{Average SR} & \boldmath{$30.00 \pm 26.46\%$} & \boldmath{$73.33 \pm 15.28\%$} & \boldmath{$83.33 \pm 11.55\%$} \\
        \cmidrule(lr){1-3} \cmidrule(lr){4-7} 
    \end{tabular*}
\end{table*}

\subsection{Experimental Setup}
Fig.~\ref{fig:experiment_setup} illustrates the experimental setup for each task. A wrist-mounted camera (Intel RealSense D435) and an external camera (Intel RealSense L515) are utilized across all tasks, with task-specific sensors incorporated for each scenario. Specifically, Task 1 incorporates a thermal camera (ThermoEye TMC256B) to acquire thermal data, while Task 2 is equipped with a tactile sensor (Meta Digit). Task 3 utilizes both sensors.

\subsubsection{Experimental datasets}
To train the VLA Executor, we collected 50 expert demonstrations for each task. The dataset consists of four components: \emph{State}, \emph{Action}, \emph{Image}, and \emph{Task prompt}.  
\begin{equation}
Dataset = \{State, Action, Image, Prompt\}
\end{equation}

\emph{State} is a 7-dimensional vector containing the robot's 6 joint angles and the gripper state, and \emph{Action} is a 7-dimensional vector comprising the 6 target joint angles and the gripper command.
\emph{Image} is collected via the wrist camera, the external camera, and additional sensors. All cameras were synchronized at a frame rate of 15 Hz; the wrist and external cameras captured $640\times480$ RGB images, while the thermal camera captured $256\times192$ thermal images. The thermal data was normalized within a standard ambient temperature range to ensure consistency, and then colorized for use. Tactile images were captured at a resolution of $640\times480$ and subsequently normalized to enhance visibility before being employed. 
Finally, the \emph{Task prompt} consists of the target task description provided in natural language.

\subsubsection{Experimental Tasks}
The proposed method was evaluated in scenarios set in industrial environments, selected from diverse real-world settings. In industrial environments, relying only on RGB camera data is often insufficient for tasks such as detecting component overheating or sorting micro-parts; therefore, such tasks require diverse sensory information, including thermal and tactile data. For example, when performing a pick-and-place operation with a heated object, thermal information becomes crucial during the detection phase. Similarly, tactile information is essential during processes such as comparing object thickness or physically interacting with the object. Motivated by these attributes, we designed three tasks where determining the object's state is difficult using a single sensory modality alone. By structuring tasks where a specific modality is highly relevant only at certain stages, we aim to verify whether the proposed model can dynamically select and utilize necessary sensory information while suppressing irrelevant inputs based on the context.

\textbf{Task 1: Battery Sorting (Visual + Thermal)} \\
The objective of this task is to discard the overheated battery while transferring the normal-temperature battery onto a conveyor belt. Since the thermal state of the batteries is indistinguishable using RGB cameras alone, the robot must selectively activate the thermal camera to accurately sort the normal battery. This task allows verification of whether the Router dynamically modulates the importance of thermal information during relevant task phases.

\textbf{Task 2: Wire Selection (Visual + Tactile)} \\
This task involves sorting two types of wires with differing thicknesses into pre-assigned bins. Specifically, thin wires must be placed at the rear, while thick wires are positioned at the front. To make visual distinction difficult, wires of identical color were selected, with a diameter difference of approximately 0.4 mm. Since this minute difference in thickness can be visually ambiguous, the robot must determine the wire's thickness using tactile feedback upon grasping. This task allows us to verify whether the Router dynamically modulates the importance of tactile information during relevant task phases.

\begin{figure*}[t] 
    \centering
    \includegraphics[width=\textwidth, keepaspectratio]{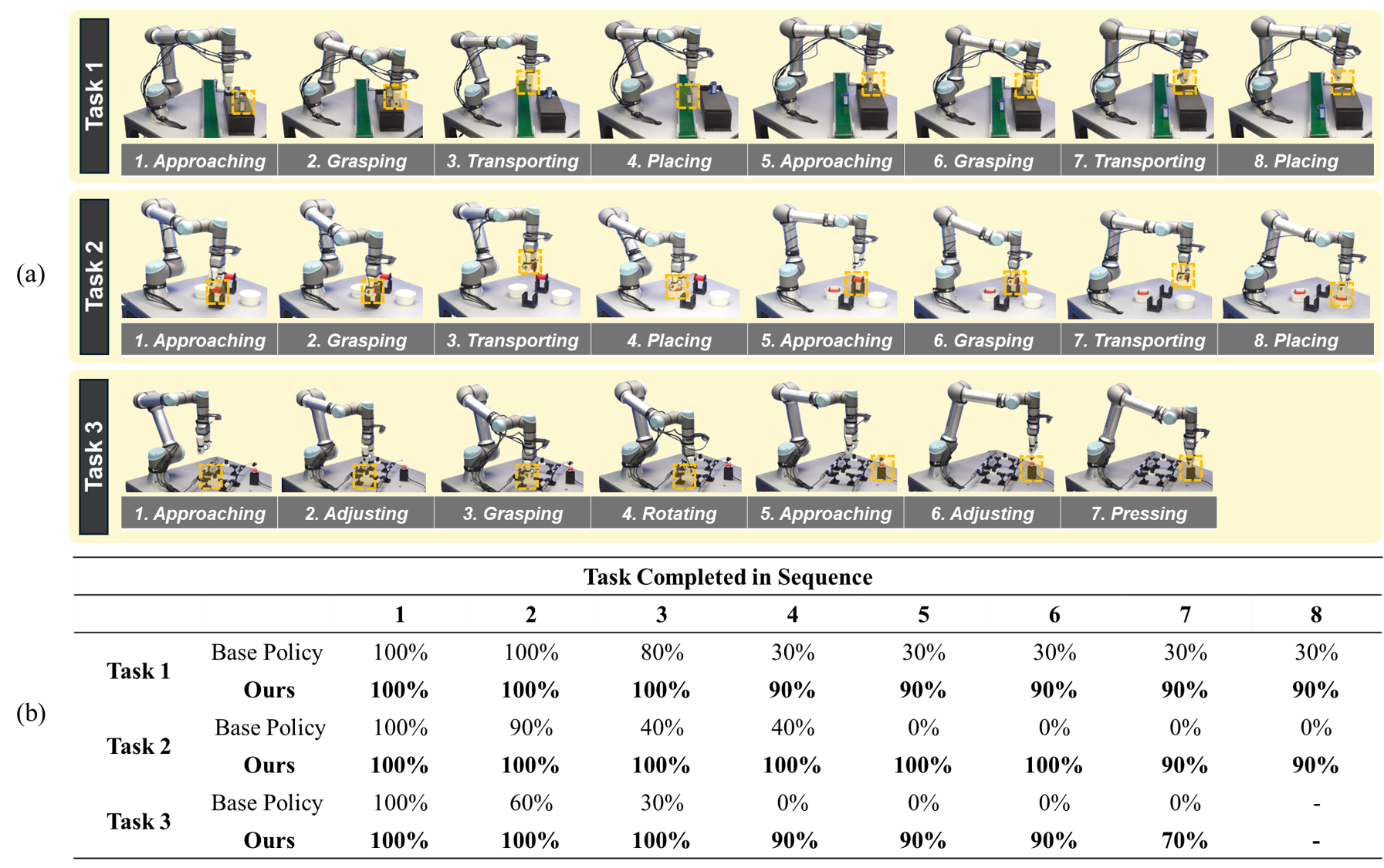}
    \caption{\textbf{Performance on Long-Horizon Sequential Tasks.} (a) Illustration of the three-stage task progression (b) Success rates for each sub-task. Our method demonstrates sustained performance across the entire sequence, highlighting its robustness in long-horizon manipulation scenarios.}
    \label{fig:long_horizon} 
\end{figure*}

\textbf{Task 3: Valve Operation (Visual + Thermal + Tactile)} \\ 
This is an emergency shutdown task simulating a pipeline overheating and leakage accident in an industrial setting. Multiple hoses and valves, along with a switch, are positioned in front of the robot. Upon detecting overheating in one of the hoses, the robot is tasked with rotating the corresponding valve to the open position, followed by pressing the switch to terminate the operation. In this scenario, the thermal camera is required for detecting overheating, while tactile information is necessary for rotating the valve. While Tasks 1 and 2 verified the selective utilization of a single auxiliary modality, Task 3 assesses the model’s capability to integrate and coordinate multiple sensory modalities.

\subsection{Analyzing Learning Efficiency with a Router}
In this paper, we propose a Dynamic Router that integrates multimodal information by assigning context-dependent weights. Table I summarizes the overall experimental results. We validate the proposed method through experiments, demonstrating its effectiveness in improving the task success rate in complex environments. The evaluation is based on the success of individual sub-tasks. Each task follows a long-horizon sequence and requires inference based on multimodal visual feedback during execution. The baseline policy using only RGB input (\textit{w/o Multimodal}) failed to achieve any success across all tasks. In contrast, incorporating multimodal feedback resulted in a success rate of 10\%. In contrast, the proposed method achieved a success rate of 73.33\%. In the following sections, we analyze these performance differences focusing on three key aspects: Router input configurations, long-horizon robustness, and the efficacy of VLM labeling.

\subsubsection{Router Input Comparison}
To identify the optimal input configuration for the router in our proposed method, we conducted comparative experiments on the following three combinations.

\begin{itemize}
  \item Wrist Camera + Prompt: The router is trained using wrist-mounted camera observations and prompts.
  \item Wrist Camera + State: The router is trained using wrist-mounted camera observations and robot states.
  \item External Camera + Prompt: The router is trained using external camera observations and prompts.
\end{itemize}

We observed that the model demonstrated superior performance when \textit{Wrist Cam + Prompt} was utilized as inputs, reaching an average of 83.33\%.
Meanwhile, the \textit{External Cam + Prompt} method achieved a relatively decent average success rate of 73.33\%; however, this represents a performance decrease of 10\% compared to the proposed method. We attribute this to the characteristics of our experimental setup, where the physical distance and perspective differences inherent to the fixed external view posed relative constraints in capturing fine-grained contacts or the precise timing of manipulation. In contrast, the \textit{Wrist Cam} provided an ego-centric view in close proximity to the target objects. This offered a distinct advantage in clearly distinguishing object state changes and task phases, particularly within the fine manipulation tasks required in this study. The \textit{Wrist Cam + State} configuration yielded a significantly lower success rate. We hypothesize that raw state information, without specific processing, acted as noise rather than a meaningful feature, thereby hindering training convergence. Notably, in Task 1, the performance of \textit{Wrist Cam + State} method even fell below that of the baseline model. This negative transfer suggests that providing inconsistent data to the router can adversely affect performance. In particular, incorrect routing decisions may lead the policy model to focus on sensor data that are irrelevant to the current task. In conclusion, the experimental results demonstrate that effective router training enhances the manipulation performance of multimodal VLA models.

\subsubsection{Long-Horizon Tasks with Dynamic Router}
Fig.~\ref{fig:long_horizon} presents the long-horizon task performance of the proposed method (\textit{w/ Router}). Prior VLA-based approaches tend to achieve relatively stable performance on short-horizon tasks; however, as the number of steps increases, their performance often degrades due to unstable skill chaining and error accumulation, as reported in prior work~\cite{yang2025lohovla,fan2025longvlaunleashinglonghorizoncapability}. Experimental results show that the baseline model exhibits a gradual decline in success rate as the number of task steps increases. In contrast, the proposed method consistently maintains a high average success rate of 83.33\% across all evaluation tasks. We attribute this improvement to the router-provided weights, which serve as an implicit step-awareness signal for the VLA model, helping it preserve task context even in long-horizon sequences.

Furthermore, the proposed approach reduces the negative impact of noise on reasoning performance by selectively filtering out irrelevant modalities. Existing models, which utilize all sensory information with equal weighting, tend to exhibit an higher failure rate as the task horizon extends, even if they successfully process the first object. By selectively attending only to task-relevant information, our method improves success rates over continuous multi-step task execution. Experimental analysis indicates that failures of the proposed method are mainly caused by object drops during transport or grasping failures, while it consistently made accurate decisions about which sensors were critical. These findings demonstrate that, in VLA training with sensory inputs, a routing mechanism that dynamically selects necessary information is more effective than naive modality fusion for high-level situational understanding and the completion of complex manipulation tasks.

\subsubsection{Validation of VLM Labeling}
\begin{figure}[t] 
    \centering
    \includegraphics[width=\columnwidth]{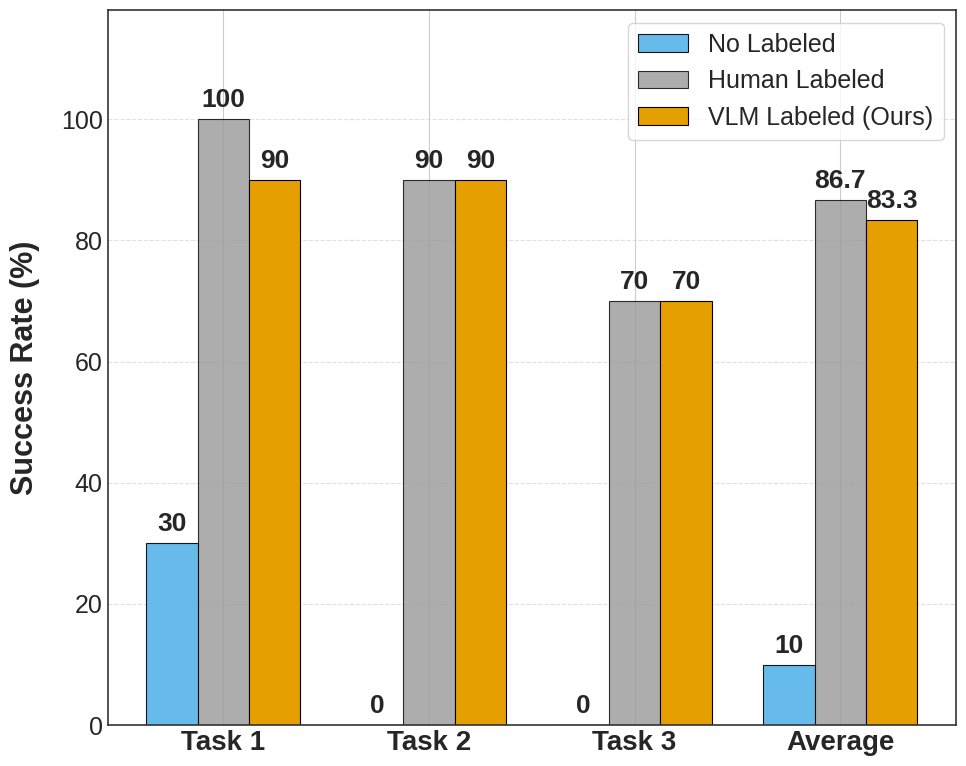}
    \caption{Performance comparison on Tasks 1--3 over 10 trials across three settings: no-labeled baseline, human-labeled, and VLM-labeled datasets.}
    \label{fig:label_compare} 
\end{figure}

In this study, we propose an automated labeling method that leverages Vision-Language Models (VLMs) capable of high-level reasoning and situational assessment. To validate our approach, we trained routing policies using VLM-generated labels and compared them against policies trained with human annotations, using task success rate as the primary evaluation metric. For each task, we conducted 10 trials and reported performance as the proportion of successful executions.

Accurate sensory labeling requires not only task understanding but also the ability to identify which sensory inputs are essential at each stage of execution. The results are summarized in Fig.~\ref{fig:label_compare}. Across all tasks, VLM-based labeling achieved an average success rate of 83.33\%, which is comparable to the 86.67\% obtained with human labeling. Across the three tasks, VLM can select task-relevant sensory information given human prompts, especially in scenarios where spatial relations among objects, temporal task structure, and contextual cues are critical.

Specifically, Tasks 1 and 2 require selecting and executing appropriate actions based on the states of two objects. Task 3 requires inferring the correct outcome (e.g., closing the appropriate valve) from environmental conditions such as the temperature of a connecting hose. These results demonstrate the practical potential of VLM-based labeling to either replace or complement human annotation. This is valuable in robot learning, where labeling costs increase rapidly as the dataset scales. In such environments, VLM-driven automated labeling provides promising approach to streamline data collection and dataset construction.

\section{Conclusion} 
In this work, we proposed a framework to enhance the efficiency and robustness of multimodal robotic manipulation. We introduced a Dynamic Router that utilizes wrist camera images and task prompts to predict real-time modality importance based on the task context. By minimizing interference from irrelevant information, our method demonstrated improved success rates in long-horizon tasks that require accurate reasoning over sensory data. Furthermore, our framework leverages a Vision-Language-Model(VLM) to determine sensory weights without the need for additional data collection or human labeling. Consequently, the resulting Vision-Language-Action (VLA) model achieved performance comparable to baselines trained with human supervision. While this study primarily focused on selectively leveraging multimodal inputs such as vision, tactile, and thermal signals, force feedback from physical interactions also offers valuable cues. Future work aims to expand the input dimensions of the Dynamic Router to incorporate these diverse modalities. We anticipate that such extensions, combined with further validation, will further enhance the robustness and generalizability of the proposed approach.

{\small
\bibliographystyle{IEEEtran}

\bibliography{references}
}

\appendices
\section*{Appendix}
\subsection{Attention Analysis for Modality Selection}
\begin{figure*}[t]
    \centering
    \includegraphics[width=0.9\textwidth]{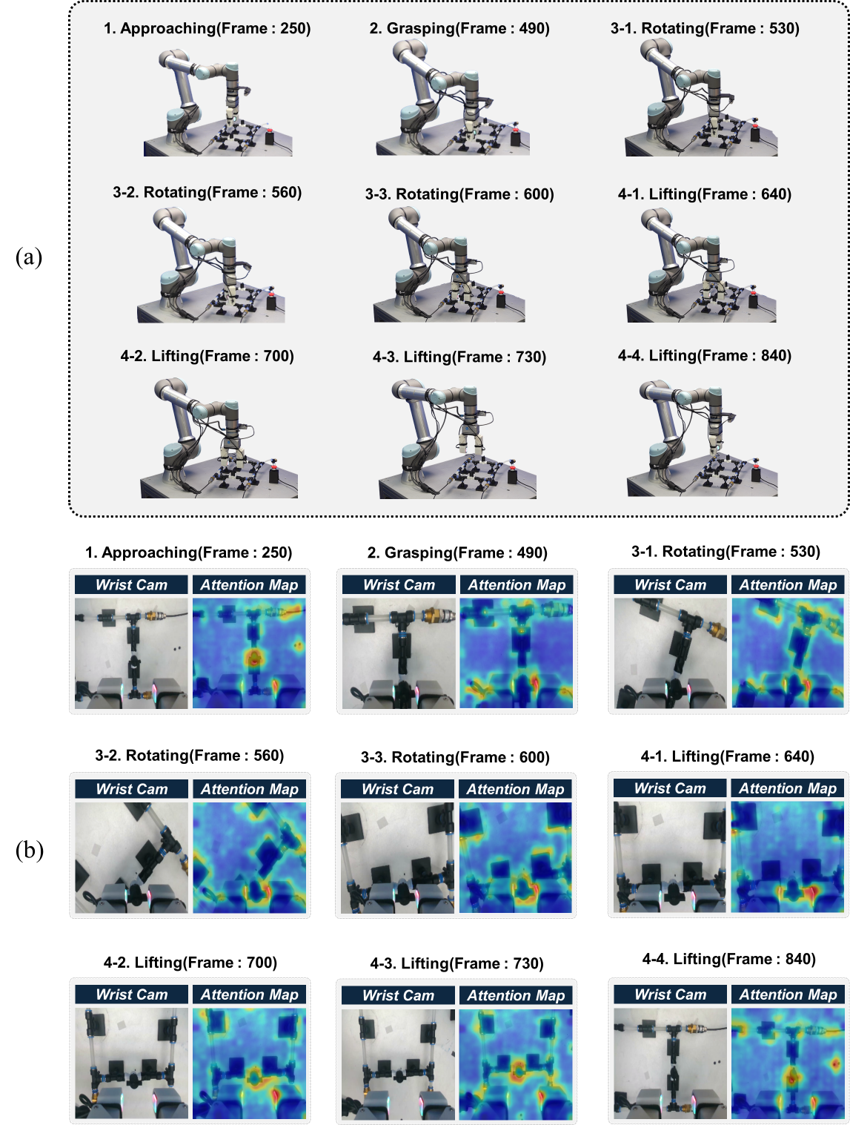}
    \caption{\textbf{Visualization of Attention Maps in the Proposed Model
} (a) Real robot manipulation sequence showing the progression of the task, including approaching, grasping, rotating, and lifting stages. (b) The router takes the wrist camera view and the prompt as inputs. We visualize the attention over the wrist camera view to illustrate how the router focuses on task-relevant regions across different stages of the manipulation task.}
    \label{fig:attention_map}
\end{figure*}

The proposed router model takes wrist camera images and prompts as inputs and outputs probability weights indicating the importance of sensory information in the current state. In this section, we visualize attention maps to analyze which regions of the visual input the router attends to during sensor selection and examine whether it associates the semantic meaning of the prompt with the physical interaction state. Fig.~\ref{fig:attention_map} compares raw wrist camera images and attention maps for the valve operation task, divided into nine key frames. The red regions in the heatmaps indicate areas of high activation.

During the initial phase where the robot approaches the valve, the router focuses on the center of the target valve, suggesting recognition of the object specified in the prompt and prioritization of visual information for path planning. When the gripper attempts to contact the valve, attention shifts to the interface between the valve and the Digit sensor mounting area, corresponding to the phase of physical contact and indicating increasing relevance of tactile sensing. During valve rotation, attention is distributed along the geometry of the handle, while in the verification stage, it spreads across the valve and surrounding regions to assess task completion. Based on these visual cues, the router dynamically adjusts weights for the external view, thermal, and tactile modalities to perform complex manipulation tasks.

\subsection{Task Definitions and Qualitative Analysis} 
\begin{figure*}[t]
    \centering
    \includegraphics[width=1.0\textwidth]{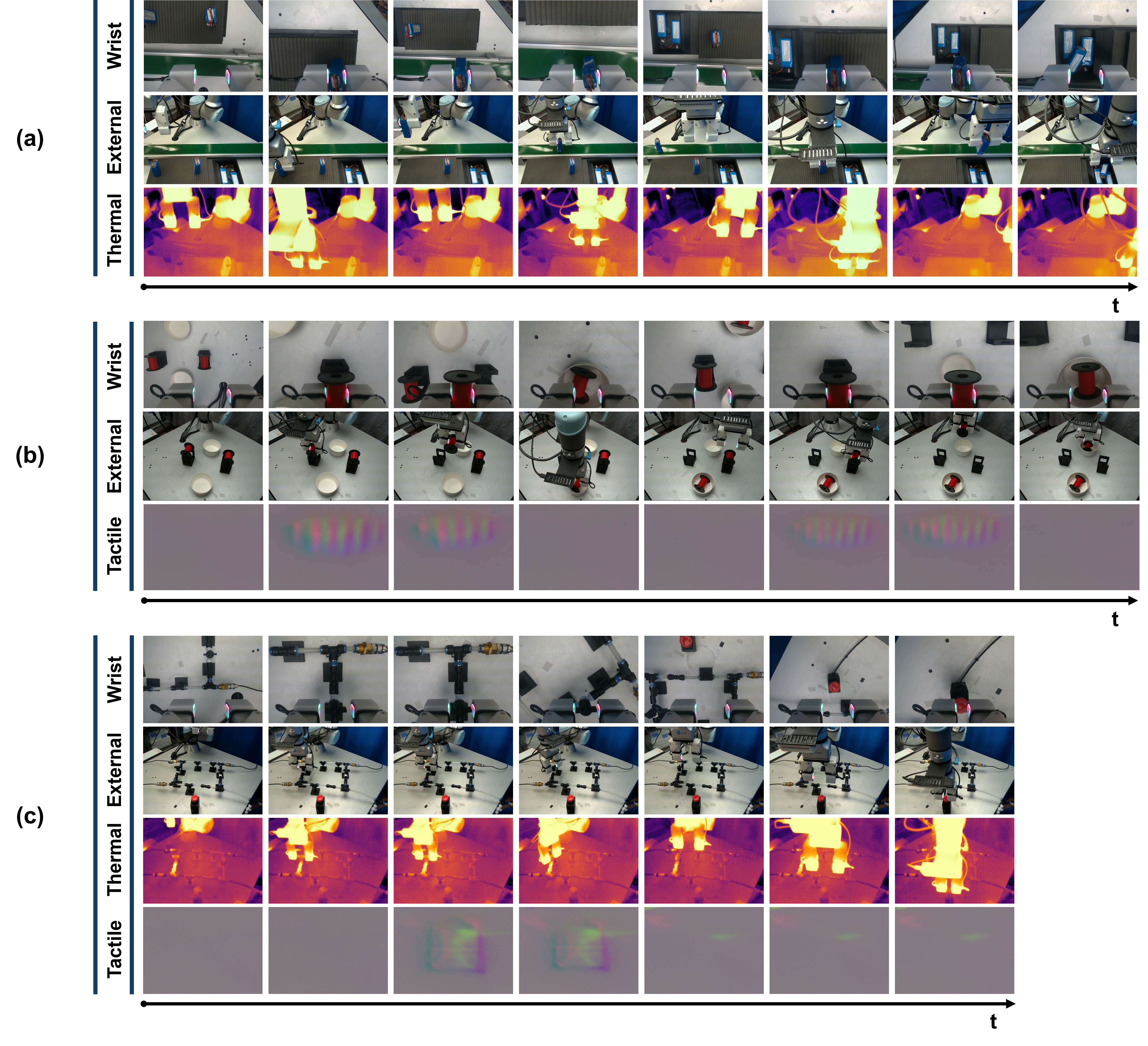}
    \caption{\textbf{Camera Views over Time for Each Task.} 
Each column represents temporal progression, while rows show sensory observations from different modalities. Task 1 (Battery Sorting), Task 2 (Wire Selection), Task 3 (Valve Operation).}
    \label{fig:time_series_task_figure1}
\end{figure*}

Fig.~\ref{fig:time_series_task_figure1} illustrates the complete execution sequences for the three experimental tasks, displaying the observations captured from each modality. This figure demonstrates how distinct sensory modalities provide unique information at each stage

\textit{Task 1: Sorting Battery (Visual + Thermal):} thermal information serves as a critical modality during the object detection and discrimination phases. While the heated battery is indistinguishable from the other battery in the wrist and external camera images, it can be reliably identified based on the thermal image. \textit{Task 2: Wire Selection (Visual + Tactile):} tactile information serves as a critical modality during the object detection and discrimination phases. While it is challenging to visually distinguish between two wires with a thickness difference of only 0.4 mm using RGB camera images, the tactile sensor exhibits distinct patterns corresponding to the wire thickness, enabling accurate comparison and identification.  \textit{Task 3: Valve Operation (Visual + Thermal + Tactile):} thermal and tactile information are both utilized as crucial modalities. The thermal camera is employed to identify which hose is generating heat in the initial phase, while tactile feedback ensures a secure grip during valve rotation.
We selected these three tasks to demonstrate that sensory information is not uniformly required throughout task execution, and that specific modalities become critical only at particular stages, as shown in Fig.~\ref{fig:time_series_task_figure1}.
Instead, specific modalities play a decisive role at specific stages. This suggests that the proposed model improves performance by selectively focusing on sensory information relevant to the current context, rather than processing all modalities uniformly.

\subsection{VLM Prompt Design}
The Vision Language Model employed in this study is Qwen3-VL-30B-A3B-Instruct. To prevent the model from drawing premature conclusions based solely on visual information, we designed the system to guide the reasoning process in a step-by-step manner. The specific reasoning process is structured into the following four stages. First, we analyze the robot's movement direction. We primarily determine whether the motion is horizontal or vertical to initially classify the task phases into those dominated by horizontal movement and those dominated by vertical movement. Second, we verify the open/closed status of the gripper. This determination is based on integrating both the visual input and the robot's actual gripper state. 
To mitigate the potential difficulty of the model in discerning fine-grained robot states solely from visual information, we explicitly injected the current gripper state into the prompt. Third, we verify the visibility of key objects, such as valves and buttons, within the current camera view. Finally, we perform a workflow consistency verification. By referencing the history of previous plans, we conduct a final verification to ensure that the preceding analysis results logically align with the predefined task sequence, before outputting the determined stage. 
\vspace{8pt}
\begin{tcolorbox}[
title=Auto Labeled VLM Prompt,
colback=gray!10,
colframe=black!60,
fonttitle=\bfseries,
breakable
]

\textbf{Role:} You are an expert who analyzes the robot's actions to activate the appropriate camera.

\textbf{Task:} \textit{Rotate the valve on the hot side and hold down the front button.}

\textbf{Gripper Action:} \textit{{Gripper Action}}

\textbf{Action History:} \textit{{History of Action}}

\textbf{Workflow Definition:} The task is divided into six phases. Each phase specifies the expected robot behavior, gripper state, movement direction, and the most informative camera modality.

\textit{Approaching phase:} The robot searches for the valve while keeping the gripper open and performing left or right movements. 

. . .

Analyze the 10 image frames and sensor data:

\begin{enumerate}
\item Movement: Is the robot moving Left/Right (Approaching/Transporting) or Up/Down (Adjusting/Lifting/Pressing)?
\item Gripper State: Is the gripper closing (Grasping), opening (Lifting), or stable?
\item Visual Cues: What objects are visible (valve, button, gripper state)?
\item Sequence Check: Strict order is Approaching $\rightarrow$ Adjusting $\rightarrow$ Grasping $\rightarrow$ Lifting $\rightarrow$ Transporting $\rightarrow$ Pressing. No skipping or reversing.
\end{enumerate}

\end{tcolorbox}

\end{document}